\documentclass[
]{ceurart}

\usepackage{listings}
\usepackage{algorithm}
\usepackage{algpseudocode}
\usepackage{multirow} 

\begin{document}



\copyrightyear{2024}
\copyrightclause{Copyright for this paper by its authors. Use permitted under Creative Commons License Attribution 4.0 International (CC BY 4.0).}

\conference{AEQUITAS 2024: Workshop on Fairness and Bias in AI | co-located with ECAI 2024, Santiago de Compostela, Spain}

\title{ProxiMix: Enhancing Fairness with Proximity 
Samples in Subgroups}




\author[1]{Jingyu Hu}[%
email=ym21669@bristol.ac.uk,
]
\address[1]{University of Bristol, Beacon House, Queens Rd, Bristol, UK}

\author[2]{Jun Hong}[%
email=jun.hong@uwe.ac.uk,
]
\address[2]{University of the West of England, Coldharbour Ln, Stoke Gifford, Bristol, UK}

\author[3]{Mengnan Du}[%
email=mengnan.du@njit.edu,
]
\address[3]{New Jersey Institute of Technology, 323 Dr Martin Luther King Jr Blvd, Newark, USA}

\author[1]{Weiru Liu}[%
email=weiru.liu@bristol.ac.uk,
]


\begin{abstract}
Many bias mitigation methods have been developed for addressing fairness issues in machine learning. We found that using linear mixup alone, a data augmentation technique, for bias mitigation, can still retain biases present in dataset labels.
Research presented in this paper aims to address this issue by proposing a novel pre-processing strategy in which both an existing mixup method and our new bias mitigation algorithm can be utilized to improve the generation of labels of augmented samples, which are proximity aware.  
Specifically, we proposed ProxiMix which keeps both pairwise and proximity relationships for fairer data augmentation. We conducted thorough experiments with three datasets, three ML models, and different hyperparameters settings. Our experimental results showed the effectiveness of ProxiMix from both fairness of predictions and fairness of recourse perspectives. 
\end{abstract}


\maketitle

\section{Introduction}

Machine learning has been used as an effective decision-making aid in more and more fields. However, concerns have been raised about the potential unjust or biased predictions by models, which can harm individual and societal values \cite{osoba2017intelligence}.
Most popular ML models are considered black-box, making it difficult to understand their internal decision-making processes. 
To address this issue, there is a growing focus on achieving fair and trustworthy ML by developing explainable and interpretable techniques \cite{burkart2021survey,arrieta2020explainable,ciatto2024symbolic}, auditing models to detect hidden bias \cite{raji2019actionable,kavouras2023fairness}, as well as mitigating the spotted bias \cite{gohar2023survey,hort2022bia}.

Among various mitigation methods, mixup-based methods have attracted increasing attention from the community.
Mixup \cite{zhang2017mixup} is a data augmentation method that linearly interpolates two samples to generate synthesized data for model generalization. 
Some research have investigate the combination of mixup with subgroup analysis for addressing fairness issues in datasets, applying it as an augmentation strategy in preprocessing \cite{navarro2023data} or a loss regularization term in training \cite{chuang2021fair}.
However, one limitation of mixup is that if the original labels in the dataset are biased, this bias can persist in the labels of mixed samples. The generated data labels can introduce additional bias to models.

To bridge the research gap, in this work, we propose ProxiMix to address the issue of biased labels in pre-processing for bias mitigation.
Motivated by the relabeling discrimination method \cite{luong2011k}, which assigns labels to instances based on their K-nearest neighbors to ensure that similar individuals have similar labels, our proposed approach adds proximity samples for re-auditing mixed labels to mitigate potential bias in mixup.
The intuition is that compared with focusing on pairwise labels, considering the labels of proximity samples as latent label relationships can reduce the probability of generating biased labels. We conducted experiments to compare the existing pairwise mixup with the proposed proximity-aware mixup on multiple models and datasets. The results showed that our ProxiMix achieves higher fairness, particularly when the original labels in the dataset are highly biased.

Our main contributions can be summarised as follows:
(1) We propose a new bias mitigation algorithm to address the label bias retainment issue in current mixup method;
(2) Subgroup preference analysis: we explore how different subgroups perform during the sampling process;
(3) Trade-off analysis: we explore the tradeoff between using our proximity-based strategy and the traditional mixup;
(4) Validation: we validate the effectiveness of our method using prediction-based metrics and the cost of counterfactual explanations from an XAI perspective.

\section{Related Work}

The fairness problem can be divided into individual and group levels. Individual fairness measures the bias by checking if similar predictions can be made for similar individuals. Group fairness compares the treatments of fairness in unprivileged and privileged groups. Fairness is achieved when the treatments are equal between groups.
Prediction-based and recourse-based fairness are two perspectives for evaluating model fairness. 
In this paper, we focus on group fairness in machine learning.

\textbf{Fairness of Prediction Outcomes}
Most fairness metrics are based on predicted outcomes. Demographic Parity (DP) \cite{zafar2017fairness} based metrics use predicted outcomes to assess whether different demographic groups are equally favored by the model. It aims for equal proportions of positive outcomes across subgroups. The DP difference between groups is called Statistical Parity Difference (SP), and DP ratio between groups is called Disparate Impact (DI). In addition to depending on predictions only, there are some fairness metrics \cite{hardt2016equality} that consider both predicted and actual outcomes. Equality of Opportunity (EO) measures the True Positive Rate (TPR) of subgroups. Equalized odds (Eodds) compares both True Positive Rate (TPR) and False Positive Rate (FPR) of each groups.

\textbf{Fairness of Recourse} Another recent research trend is to apply Explainable Artificial Intelligence (XAI) methods to address fairness issues. One of the key components in this area is counterfactual explanation (CE), sometimes also called as algorithm recourse. 
CE focuses on explaining why a particular outcome occurred instead of an alternative plausible outcome. \cite{buchanan1984rule,gregor1999explanations}. Recourse refers to identifying the closest counterfactuals that could alter the result with minimal feature changes. Several algorithms have been developed to generate such counterfactual explanations for machine learning models \cite{mothilal2020explaining,wachter2017counterfactual,brughmans2023nice}. 
The concept of fairness of recourse are proposed by \cite{gupta2019equalizing} and defined as the disparity of the mean cost to achieve the desirable recourse among the unprivileged subgroups. \cite{kavouras2023fairness,artelt2022explain} proposed metrics based on the cost of counterfactual explanation to measure fairness performance across subgroups. Predictive Counterfactual Fairness (PreCoF) \cite{goethals2023precof} utilises CEs to detect underlying patterns for the discrimination in the model.

\textbf{Bias Mitigation Methods}
Bias mitigation methods can be categorized into three stages: pre-processing, in-processing, and post-processing \cite{hort2022bia,friedler2019comparative}. 
Pre processing mitigations aim to reduce bias by modifying and creating a fairer training dataset \cite{kamiran2009classifying,feldman2015certifying,sun2022towards}. In-processing mitigation occurs during training by adding regularization and constraints to models \cite{chuang2021fair,kamiran2010discrimination}. Mitigations in the post-processing stage like calibration are applied after a model has been successfully trained \cite{artelt2022explain,pleiss2017fairness}. 
Both pre-processing and post-processing-based methods are model-agnostic as they occur before and after the model training.

Over-sampling in the pre-processing stage refers to changing the distribution of the training dataset by adding more samples. Duplicating instances of the unprivileged group is one straightforward strategy \cite{amend2021improving,morano2020bias}. \cite{dablain2022towards,biasWhyHow222} generate synthetic samples around the unprivileged group to mitigate bias. MixSG \cite{navarro2023data} takes both the privileged and unprivileged groups into consideration when synthesizing new data using mixup, but the potential bias in generated labels has not been discussed yet.

\section{Preliminaries and Problem Statement}
\textbf{Notations}
Given the dataset ${D}=\left\{\left(X, Y, Z\right)\right\}_{i=1}^N$ with $N$ samples, where $X$ is a set of feature space, and each feature $x$ in $X$ has a set of values in $d_{x_i}$, label $Y \in \mathcal{Y}:=\{0,1\}$, and a sensitive attribute $Z \in \mathcal{Z}:=\{0,1\}$. The dataset is divided into training set $D_{train}$ and test set $D_{test}$. We use $D_{train}$ to fit a classifier model $ f: \mathcal{X} \rightarrow \mathcal{Y}$ and $D_{test}$ to assess the model's prediction and fairness performance.
Fairness is measured by the model's performance on the difference between subgroups identified by $Z$. We define the unprivileged/minority group when Z=0, and Z=1 is the privileged/majority group.

\textbf{Mixup Strategy in Fairness}
Mixup \cite{zhang2017mixup} is a data augmentation technique that involves blending pairs of samples to create new synthetic training examples.
The premise of mixup is that linear combinations of features will result in the same linear combinations of target labels. Thus, mixup applies stochastic linear combinations to samples $S_0(x_0, y_0)$, $S_1(x_1, y_1)$ to generate a new sample $\tilde{S}(\tilde{x}, \tilde{y})$, with random parameters $\lambda$ drawn from a Beta distribution.

\begin{align}
\tilde{x}=\lambda * x_0+(1-\lambda) * x_1, \quad \text { where } x_0, x_1 \text { are input vectors } \\
\tilde{y}=\lambda * y_0+(1-\lambda) * y_1, \quad \text { where } y_0, y_1 \text { are target labels }
\end{align}

To address fairness concerns, previous research has explored the practice of sampling $S_0$ and $S_1$ from different subgroups, applying this step to both pre-processing stage like mixSG \cite{navarro2023data} and in-processing stage like fairMixup \cite{chuang2021fair} as bias mitigation methods.

\label{sec:mixup-challenge}
\textbf{Bias Persist After Mixup}
The premise of mixup lies in the linear relationship between features and labels.
The challenge here is if the original labels in the dataset are biased, the labels of mixed samples can retain this bias. The newly generated biased samples can impact the fairness of the trained model.

Here is a toy example. Table \ref{tab:sample_demo} presents simplified instances of individual income predictions by the ML model. 
The predicted label $Y$ indicates whether an individual is high-income ($>50K$) or low-income ($\leq 50K$). The features $X$ used for prediction include $Age$, $Occupation$, $Gender$, $Capital Gain$, and $Capital Loss$. $Gende$r is considered as the sensitive attribute $Z$, dividing the data into subgroups. Here, we consider the female subgroup as unprivileged. 

The table shows individual features of male samples ($M1$ and $M2$) and the female sample ($F2$) are remarkably similar (Officer with similar $Capital Gain$ and $Age$), but with different income labels. This shows initial bias that female and male groups are treated unequally.

\begin{table}[t]
    \centering
    \caption{Simplified Sample Examples of Individual Income Prediction}
    \includegraphics[width=0.85\linewidth]{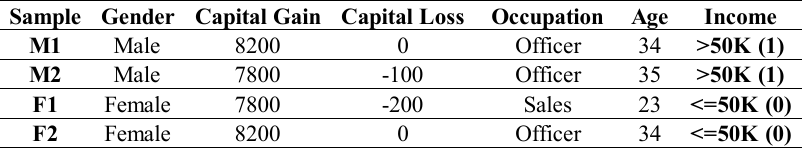}
    \label{tab:sample_demo}
\end{table}

We follow the mixSG method to select one sample from one subgroup and another from the other subgroup to generate ($\tilde{x},\tilde{y}$). Assume we have chosen one sample $F2$ from the female subgroup, $F2$ will be randomly paired with either $M1$ or $M2$ from the male subgroup. If the mixture ratio $\lambda$ of the female sample $F2$ is over 50\%, we say the mixed sample $S_{FM}$ is female. Otherwise, $S_{FM}$ is male.

When the random $\lambda$ = 0.8, $S_{FM}$ will be a female sample. And the label $y_{FM}$ of the mixed female sample will primarily depend on the label from female $F2$, meaning that both combinations of $F2$ with $M1$ or $M2$ will have a high probability of low income ($\leq 50K$). 
Though individual features of high-income men (M1 and M2) and low-income women (F2) are remarkably similar (Officer with similar capital gain and age), mixed label still indicates a tendency toward lower incomes for female.
If $\lambda$ = 0.2, the mixed sample will be most depend on the label from the male sample and the generated sample becomes male with high income. 
The labels of mixed samples are heavily influenced by gender.
Considering the initial bias in the dataset, new samples generated by mixup can deepen gender bias against unprivileged groups, causing the model to be more likely to predict male samples as high-income and female samples as low-income under similar conditions.

\section{Methodology and Experiment Design}

To address the issue of possible biased label for mix-up, we proposed a method called ProxiMix for improvments.
It synthesizes ${D_{\text{new}}}=\left\{\left(X, Y, Z\right)\right\}_{j=1}^{K}$ from $D_{train}$ with the consideration of both pairwise and proximity samples, to reduce dataset bias. 
Fitting the model with fairer dataset ${D}^{\prime}_{train} ={D}_{train} \cup {D_{\text{new}}}$ is expected to improve its fairness performance.

\subsection{ProxiMix Algorithm}

\begin{figure}[t]
    \centering
    \includegraphics[width=0.95\linewidth]{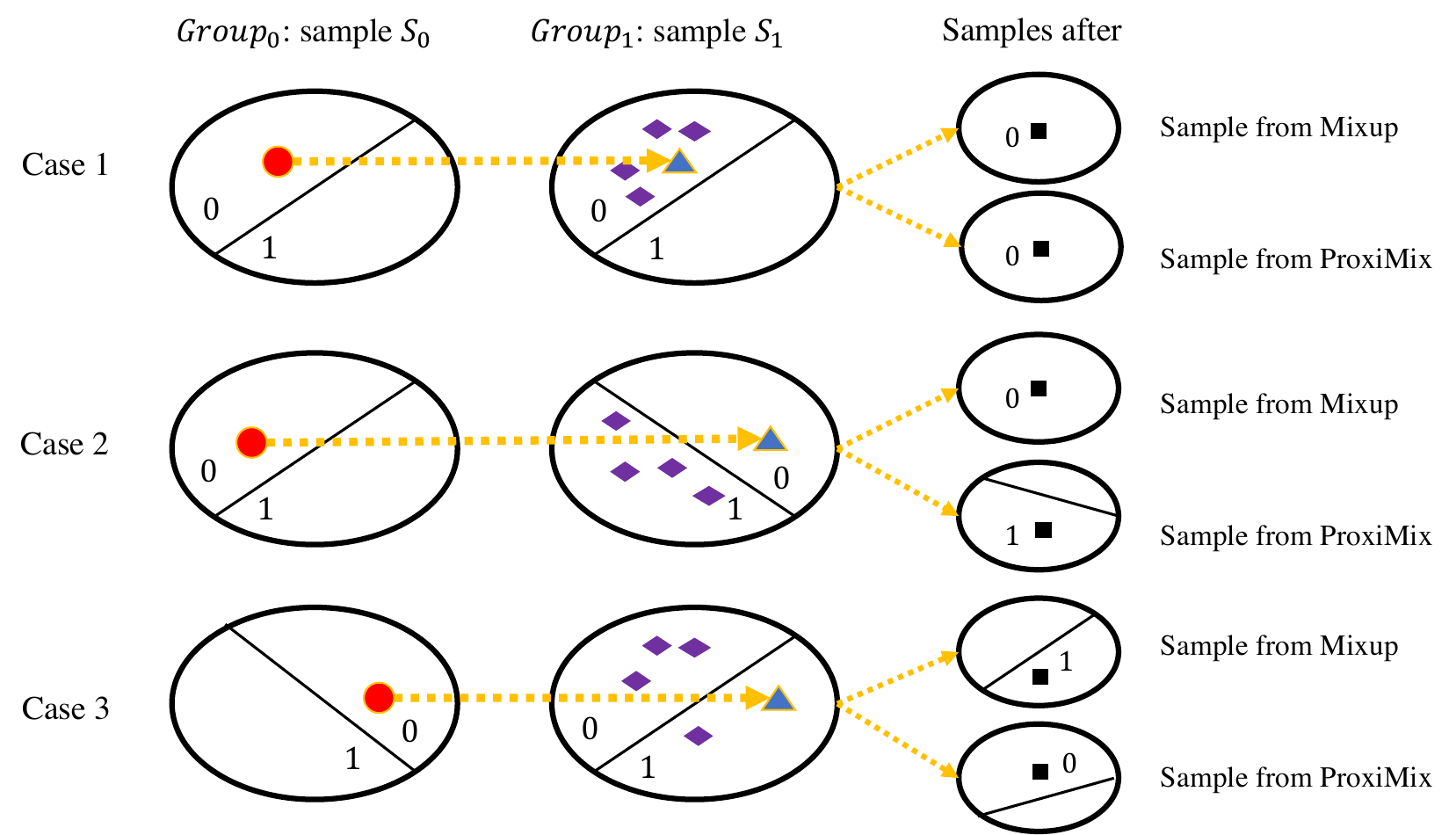}
    \caption{A comparison between proximity-based mixup and linear mixup. The red circle represents $S_0$, the blue triangle represents $S_1$, the purple diamonds represent the proximity set $D_p$, and the black square indicates the samples after mixing up. Here, we consider the particular case for case three where labels of most proximity samples are opposite to $S_1$. The mixing ratio is set to $0.5$.}
    \label{fig:how-works}
\end{figure}

\textbf{The Importance of Proximity Awareness}
Given a sample $S_0$ from group $D_{\text{train}}(Z=0)$, and another sample $S_1$ from $D_{\text{train}}(Z=1)$, the proximity samples set of $S_1$ is defined as $D_p=\{S_{p_0},S_{p_1},...,S_{p_m}\}$. The label value of each sample can be either $0$ or $1$. 
We illustrate three cases when mixing up two samples $S_0$ and $S_1$:
\textbf{(1) Case 1:} Labels of $S_0$, $S_1$ and all of their proximity samples are the same.
\textbf{(2) Case 2:} Labels of $S_0$ and $S_1$ are the same, but there exist different labels among proximity samples $D_p$.
\textbf{(3) Case 3:} Labels of $S_0$ and $S_1$ are different.
Figure \ref{fig:how-works} presents these three cases.

In Case 1, linear mixing and proximity yield the same results because there are no impurities between the two samples.
In Case 2, both samples $S_0$ and $S_1$ have the same label. This implies that direct mixing will result in all labels becoming $0$ regardless of the mixing ratio. This approach ignores the samples from $1$ in between and can potentially introduce bias when predicting subgroups with the $1$ label.
In Case 3, the mixed label depends on the mixing rate when using mixup directly. Specifically, the mixed label becomes $1$ when the mixing rate exceeds 0.5. However, we can see in the example that the majority of the proximity samples $D_p$ between $0$ and $1$ belong to $0$. It suggests that the probability of being classified as $0$ should be higher. Considering the proportion of proximity labels can enhance the probability of being classified as $0$.

\newpage
\textbf{ProxiMix Algorithm Design}
ProxiMix consists of two parts: we first introduce proximity-based mixed label $Y_{sim}$ and then combine $Y_{sim}$ with $Y_{\lambda}$ from the existing mixup \cite{navarro2023data} using d-adjusted balancing degree.

As discussed above, the current mixup approach does not account for potential biases in labels.
Our proposal aims to determine the mixed label by considering the proportions of labels in proximity samples. 
Specifically, when mixing two samples, $S_0$ and $S_1$, we calculate their Euclidean distance with their one-hot encoded features\footnote{\url{scikit-learn.org/stable/modules/generated/sklearn.preprocessing.OneHotEncoder.html}}, denoted as $P_{dis} = ||x_0 - x_1||$, to measure their proximity. 
Then, we select all the samples that are within the $P_{dis}$ distance from $S_0$ to form a potential proximity samples set ProxiSet. The final mixed label for $S_0$ and $S_1$ is assigned based on the label with the larger proportion within the $S_0 \cup ProxiSet$.

Let's look back at the toy example: $NewSet = \{F2, M1, M2$\} when we want to mix $F2$ with either $M1$ or $M2$. Two-thirds of the labels in the $NewSet$ is high income, so that the proximity-based mixed $Y_{sim}$ is high income.

We combine our proximity-based $Y_{Sim}$ with $Y_{\lambda}$ from the current mixup to form the new definition of mixed $\tilde{Y}$, achieved by calculating $d*Y_{\lambda} + (1 - d) *Y_{Sim}$, where $d$ is a balancing degree between 0 and 1. The algorithm pseudocode is described in Algorithm 1.

\begin{algorithm}[ht]
\caption{ProxiMix Algorithm}\label{mixed_label_algo}
\begin{algorithmic}
\State \textbf{Input} $S_0(x_0,y_0,z_0) \sim D_{\text{train}}(Z=0), S_1(x_1,y_1,z_1) \sim D_{\text{train}}(Z=1)$
\Procedure{ProxiMix}{$S_0, S_1, D_{\text{train}},d$}
\Procedure{Proximity-Based-Mixed}{$S_0, S_1, D_{\text{train}}$}
\State $ProxiSet = [ ]$ .
\State $P_{dis} = ||x_0 - x_1||$
\For{each sample $S_{i}(x_i,y_i,z_i)$ in $D_{\text{train}}(Z=1)$}
    \State $P_{cur} = ||x_i - x_0||$

    \If {$P_{cur} \leq P_{dis}$}  
        \State Add $S_{i}$ to $ProxiSet$.
    \EndIf
\EndFor
\State $NewSet=S_0 \cup ProxiSet$
\State $Y_{Sim} = Label\_Counts (Y \in NewSet) /size(NewSet)$

\EndProcedure
\Procedure{Lambda-Based-Mix}{$S_0, S_1$}
\State $\lambda = \operatorname{Beta}(\alpha, \alpha)$
\State $Y_{\lambda}=\lambda * y_0+(1-\lambda) * y_1$
\EndProcedure
\State $\tilde{Y}= d*Y_{\lambda} + (1 - d) *Y_{Sim}, d \in [0,1]$
\State \textbf{Return} $\tilde{Y}$
\EndProcedure
\end{algorithmic}
\end{algorithm}

\begin{figure}[t]
    \centering
    \includegraphics[width=1\linewidth]{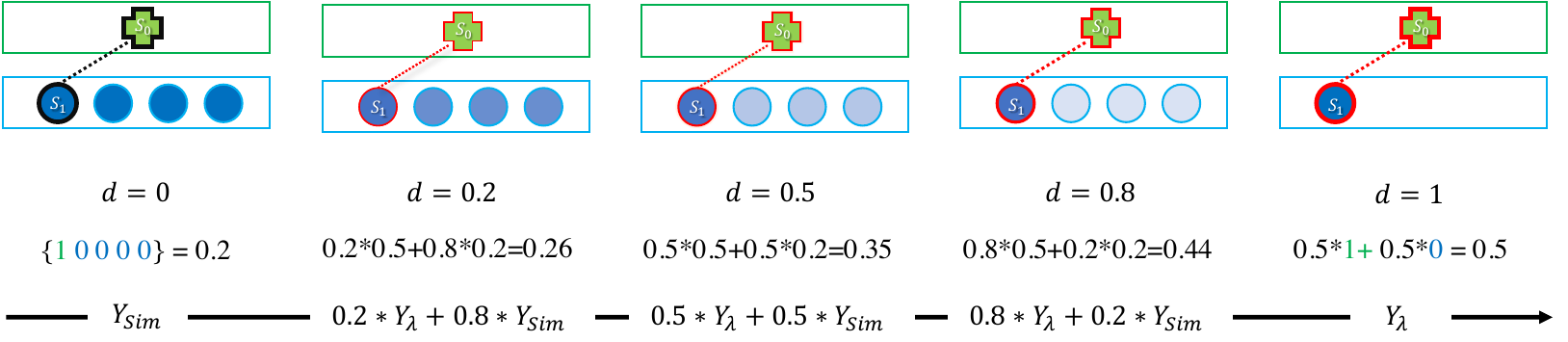}
    \caption{An Example of ProxiMix with Balancing $d=[1,0.8,0.5,0.2,0], \lambda=0.5$}
    \label{fig:eg-proximix}
\end{figure}

\newpage
Fig. \ref{fig:eg-proximix} shows an example of how ProxiMix works. Samples are categorized into two subgroups, green and blue, based on their colors. The shape of each sample represents its label: circles for label 0, and plus-signs for label 1. Specifically, the green circle ($S_0$) and the blue plus-sign ($S_1$) are two samples selected for ProxiMix. 
The new label of the mixed samples changes with different values of the balancing parameter $d$. The varying shades of blue samples represent the impact degree of $Y_{sim}$, while the thickness of the red lines between $S_0$ and $S_1$ represents the strength of $Y_\lambda$. The black line indicates no consideration for $Y_{\lambda}$.
For $d=1$, it employs the original mixup $Y_{\lambda}$; for $d=0$, it utilizes our proximity-based $Y_{Sim}$ exclusively; and it combines the two for values in between.
We will discuss how different $d$ impact the model performance in Section \ref{sec:diff_d}.

\textbf{Accelerating Calculation of ProxiMix in Practice}
Our core idea is to introduce proximity samples' label set $ProxiSet$ as a reference when performing label mixup. 
To enhance computational efficiency, we find $ProxiSet$ first in practice.
Our implementation is as follows: (1) Given a randomly selected sample $S_0$ from $D_{train}(Z=z)$, we first find its $ProxiSet$ from $D_{train}(Z=\neg{z})$. {$ProxiSet$ contains $K$ samples that are proximal to $S_0$; (2) Then, we treat each sample in $ProxiSet$ as $S_1$ and sequentially mix it with $S_0$, following the `furthest-first' rule. It means the mixing begins with the sample in $ProxiSet$ that is furthest from $S_0$. After each mix, we remove the used sample from $ProxiSet$; (3) Repeat this process $K/M$ times until the desired $M$ new samples are generated.
The generated samples are merged to $D_{train}$ as training samples for classification model.

\subsection{Experiment Setting}
\label{sec:expe_equation}
Fig. \ref{fig:workflow} presents the overall workflow of our experiment. The parameter balancing degree $d$ in our mixup algorithm is tested with values ranging from 0 to 1, in steps of 0.1. The proximity samples for each round are set to 25. we consider proximity when there are at least 5 neighbors to ensure credibility. The mixing ratio $\lambda$ is randomly generated from the Beta(1,1) distribution.

\begin{figure}[ht]
    \centering
    \includegraphics[width=0.7\linewidth]{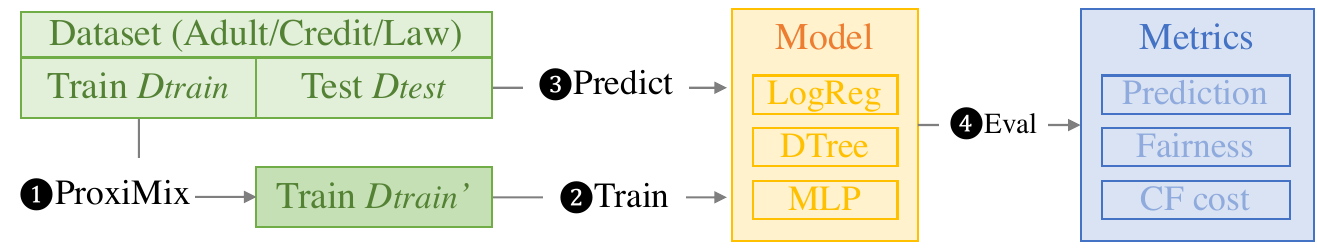}
    \caption{The Experiment Workflow}
    \label{fig:workflow}
\end{figure}

\textbf{Datasets}
The experiment is conducted on three datasets for classification problems: 
(1) Adult income dataset \cite{adultdsref}: predicting whether a person's annual income exceeds 50K (high/low-income);
(2) Law school dataset \cite{lawschoolref}: predicting whether a person's in law school will fail/pass the exam;
(3) Credit default dataset \cite{creditdsref}: predicting whether a person's credit payment will be on-time/overdue.

\textbf{Models} Three models including logistic regression (LogReg), decision trees (DT) and multi-layer perceptron (MLP) are tested.
All implementations are based on scikit-learn\footnote{\url{scikit-learn.org/}}. The maximum depth is 7 in the decision tree. We use a three-layer MLP with 128 neurons in the ith hidden layer, `rule' as the activation function, and a maximum of 1500 iterations. The random seed is set to 42 for reproducible results.

\textbf{Metrics}
Prediction performance metrics are based on True Positive (TP), False Positive (FP),False Negative (FN),True Negative (TN) in the confusion matrix.
The equations of Precision, Recall, and F1-score are as follows. Recall is also called True Positive Rate.

\begin{equation}
Precision = \frac{TP}{TP + FP}; \\
Recall = \frac{TP}{TP + FN}; \\
F1-score = \frac{2 \cdot Precision \cdot Recall}{Precision + Recall}
\end{equation}

The following equations calculate the True Positive Rate (TPR) and False Positive Rate (FPR) in the subgroup where the sensitive attribute $Z=z$.

\begin{equation}
TPR_{z} = \frac{TP}{TP + FN}; \\
FPR_{z} = \frac{FP}{FP + TN} where (D(Z=z))
\end{equation}

Fairness performance is evaluated by Demographic Parity (DP) and Equalized Odds (Eodds) between subgroups\footnote{\url{fairlearn.org/}}. 
We define the label for the unprivileged group as $z_0$ and for the privileged group as $z_1$.
Their difference and ratio between $DP_{z_0}$ and $DP_{z_1}$ are noted as $\Delta DP, DP\%$.

\begin{equation}
DP_{z_0}=\operatorname{P}\left(f(x)=1 \mid Z=z_0\right); DP_{Z_1}=\operatorname{P}\left(f(x)=1 \mid Z=z_1\right)\\
\end{equation}

\begin{equation}
\Delta DP= DP_{z_1}-DP_{z_0}; \\
DP\%=\frac{DP_{z_0}}{DP_{z_1}}
\end{equation}

Eodds difference $\Delta Eodds$ is defined as the greater one of $TPR$ and $FPR$ across subgroups, and eodds ratio $Eodds\%$ is the smaller metrics of TPR and FPR ratio.

\begin{equation}
\label{alg:eod}
\Delta Eodds=Max(TPR_{z_1}-TPR_{z_0}, FPR_{z_1}-FPR_{z_0})
\end{equation}

\begin{equation}
\label{alg:eor}
Eodds \%=Min \left(\frac{TPR_{z_0}}{TPR_{z_1}}, \frac{FPR_{z_0}}{FPR_{z_1}}\right)
\end{equation}

Counterfactual explanation cost \cite{artelt2022explain} is also assessed across subgroups to examine fairness from an XAI perspective.
Given a classification model $f$, the counterfactual explanation of a sample $d_s \in D$ can be denoted as $d_{cf}=CF(d_s, f)$.
The cost of a counterfactual explanation is the distance between $d_{s}$ and $d_{cf}$. In this way, we can compute the counterfactual cost for each sample in dataset $D$.
The average costs of counterfactuals across different groups can be considered as a measure of fairness: with the cost gap between groups (e.g., females and males) increasing, the model's unfairness also grows.
Our evaluation follows the implementation of counterfactual explanation cost package\footnote{\url{github.com/HammerLabML/ModelAgnosticGroupFairnessCounterfactuals/}}, and specifically, we opt counterfactual explanations cost without constraints as metrics.

\section{Results}
In section \ref{sec:diff-com}, we fix the balancing degree $d$ of ProxiMix and examined the impact of different sampling modes for subgroups on the outcomes.
In section \ref{sec:diff_d}, we fix the sampling mode and explored the impact of different balancing degrees $d$ on the results.
To ensure the consistency of findings, Section \ref{sec:res_cf} assesses the effectiveness of ProxiMix from counterfactual cost perspective.

\subsection{Sampling Mode Preferences in ProxiMix with Fixed Balancing Degree}
\label{sec:diff-com}

ProxiMix is built on the mixup concept, which involves continuously selecting and mixing two samples to generate new data. To identify which combinations of samples had a more positive impact on the model's performance, we divide the dataset into different subgroups and sample from them. 

\begin{table*}[t]
    \centering
    \caption{Four different subgroup combinations for sampling in Adult, Law, Credit datasets}
    \label{tab:sample-groups}
    \resizebox{0.82\textwidth}{!}{
    \begin{tabular}{cccccc}
        \hline
         & C1 (z,y) & C2 (z,y) & C3 (z,y) & C4 (z,y) \\
        \hline
        Adult & female, low-income & female, high-income & male, low-income & male, high-income \\
        Law & female, failed & female, passed & male, failed & male, passed \\
        Credit & female, on-time & female, overdue & male, on-time & male, overdue \\
        \hline
         & C1'($\bar{y}$) & C1'($\bar{y}$) & C3'($\bar{y}$) & C3'($\bar{y}$) \\
        \hline
        Adult & male group & male group & female group & female group \\
        Law  & male group & male group & female group & female group \\
        Credit & male group & male group & female group & female group \\
        \hline
    \end{tabular}
    }
\end{table*}

There are four subgroups with considerations on both labels and values of a single sensitive feature in $Z$. The first sample selected from each group $D_{train}(Y=y,Z=z)$ is notated as $C1$, $C3$, the second sample selected from the subgroup $D_{train}(Y=\bar{y})$ which has the opposite sensitive label is notated as $C1'$,$C2'$,$C3'$,$C4'$, respectively. 
In Table \ref{tab:sample-groups}, $C1$ is sampled from $<$female, low-income$>$ subgroup in the Adult dataset, from the $<$female, failed$>$ subgroup in Law dataset, and from the $<$female, on-time$>$ subgroup from the Credit dataset respectively.
$C1'$ refers to the sample selected from the male group in the adult, law and credit datasets. All sampling combinations are listed in Table \ref{tab:sample-groups}. We denote the sample derived from  ProxiMix with different sampling combination modes as $C_i \odot C_j$, where $C_i \in \{C1,C2,C3,C4 \}, C_j \in \{C1',C3'\}$.

Table \ref{tab:res-diff-combinations2} presents models performance using ProxiMix under four sampling combinations $C_i \odot C_j$ and compares it with performance without any augmentation (baseline).

In the adult dataset, we found that different subgroup sampling combinations have different impacts on ProxiMix performance. The $C2 \odot C1'$ (augmenting high-income female) significantly improves the fairness performance of both decision tree and logistic regression models. In contrast, $C1 \odot C1'$(augmenting low-income female) degrades the fairness of both models, suggesting it introduces extra bias to the underrepresented group.
This implies that focusing on underrepresented labels in the unprivileged group when generating samples (such as high income) can greatly improve fairness performance.

In the Law dataset, nearly all mixup methods enhance model prediction performance, but only marginally improve fairness. This is because fairness performance DP\% without any augmentation already exceeds 90\%, indicating the minimal bias in the model. Therefore, the improvement potential is limited. 

Overall, ProxiMix enhances fairness when a model displays significant bias. Also, the choice of the subgroup for sampling during mixup is important: some enhance fairness, while others can even worsen it. 

\begin{table*}[t]
\centering
\caption{Prediction (F1 score) and Fairness (DP\%) Performance Comparison across Different Sampling Subgroups in Adult and Law School Datasets ($d$=0.5, LogReg stands for logistic regression, and DT represents the decision tree.}
\label{tab:res-diff-combinations2}
\setlength{\tabcolsep}{3mm}
\resizebox{\linewidth}{!}{
\begin{tabular}{ccccccccc}
\hline
Dataset & \multicolumn{4}{c}{Adult Income}      & \multicolumn{4}{c}{Law School} \\ 
\hline
Model & \multicolumn{2}{c}{LogReg} & \multicolumn{2}{c}{DT} & \multicolumn{2}{c}{LogReg} & \multicolumn{2}{c}{DT} \\
\hline
 & F1 Score & DP\%   & F1 Score & DP\%   & F1 Score   & DP\%    & F1 Score  & DP\%    \\
\hline
Baseline    & 0.7791  & 0.2892 & 0.7782  & 0.2847 & 0.6408  & \underline{0.9856} & 0.6146  & \textbf{0.9935} \\
$C1 \odot C1' $    & 0.7758  & 0.2439 & 0.7749  & 0.2792 & 0.6680  & 0.9261 & 0.6336  & 0.9831 \\
$C2 \odot C1' $       & 0.7820  & \textbf{0.4730} & 0.7729  & \textbf{0.3698} & 0.6279  & \textbf{0.9948} & 0.6428  & \underline{0.9925} \\
$C3 \odot C3' $       & 0.7705  & 0.2625 & 0.7721  & 0.2971 & 0.6696  & 0.9619 & 0.6309  & 0.9837 \\
$C4 \odot C3' $       & 0.7884  & \underline{0.2889} & 0.7780  & \underline{0.2988} & 0.6251  & 0.9840 & 0.6369  & 0.9921 \\
\hline
\end{tabular}
}
\small
\end{table*}

\subsection{The Impact of Balancing Degree in ProxiMix}
\label{sec:diff_d}

The above section discussed the different sampling strategies with a balanced mixup ($d=0.5$). This section explores how different $d$ in ProxiMix can impact model performance. Here, we fix strategy $C_i \odot C_j$ while changing balance degree $d$.

Fig. \ref{fig:mix-credit-res} illustrates the impact of data augmentation on model fairness in the Credit dataset, under $C1 \odot C1' $ and $C3 \odot C3'$ strategies, with different degree $d$. The trend shows most combinations positively affect a model fairness, with an optimal $d$ that maximizes fairness improvements. The best performance is achieved at \textit{d=0.7} for the $C1 \odot C1'$ strategy, while for $C3 \odot C3'$, the optimal performance is reached at \textit{d=0.2}.

\begin{figure}[t]
    \centering
    \includegraphics[width=0.95\linewidth]{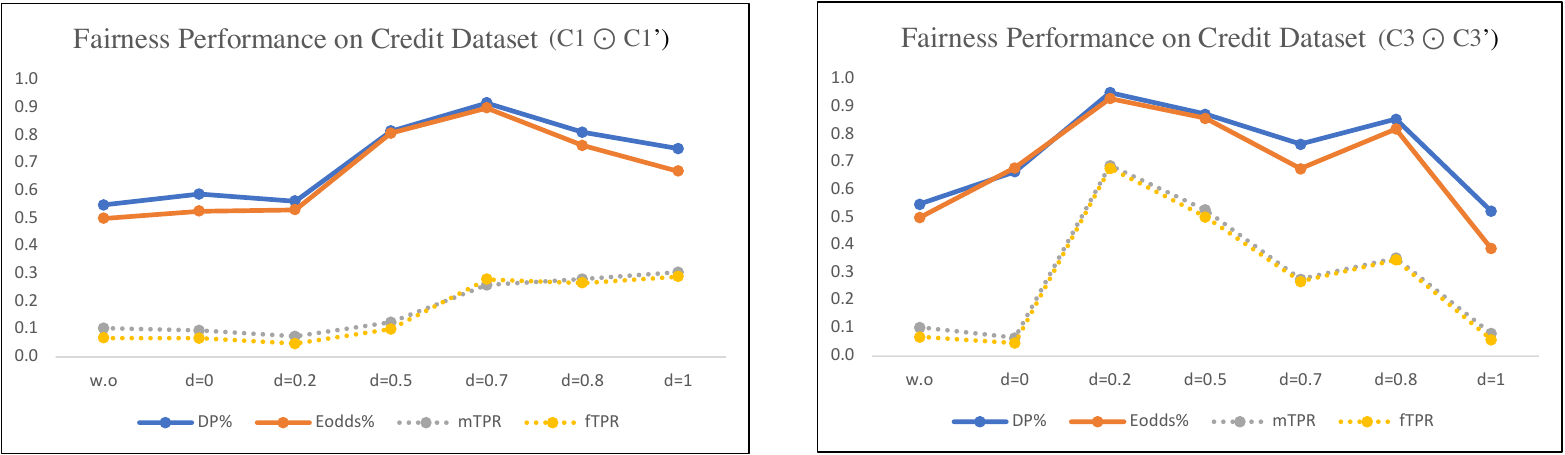}
    \caption{The fairness performance changes under different balancing degree $d$ in Credit Default dataset under MLP model (fTPR: TPR in female group, mTPF: TPR in male group, $d = [0, 0.2, 0.5, 0.7, 0.8, 1]$, refer to Appendix \ref{app:credit_appendix} for detailed results) }
    \label{fig:mix-credit-res}
\end{figure}

Similar patterns are observed in the adult dataset (Fig.\ref{fig:mixed-demo-adult}): the impact of different values of $d$ on the model also shows a trend. Specifically, data generated with the $C2 \odot C2'$ strategy shows the better improvement in model fairness when $d$ ranges from \textit{0.2} to \textit{0.5}.

\begin{figure}[t]
    \centering
    \includegraphics[width=0.95\linewidth]{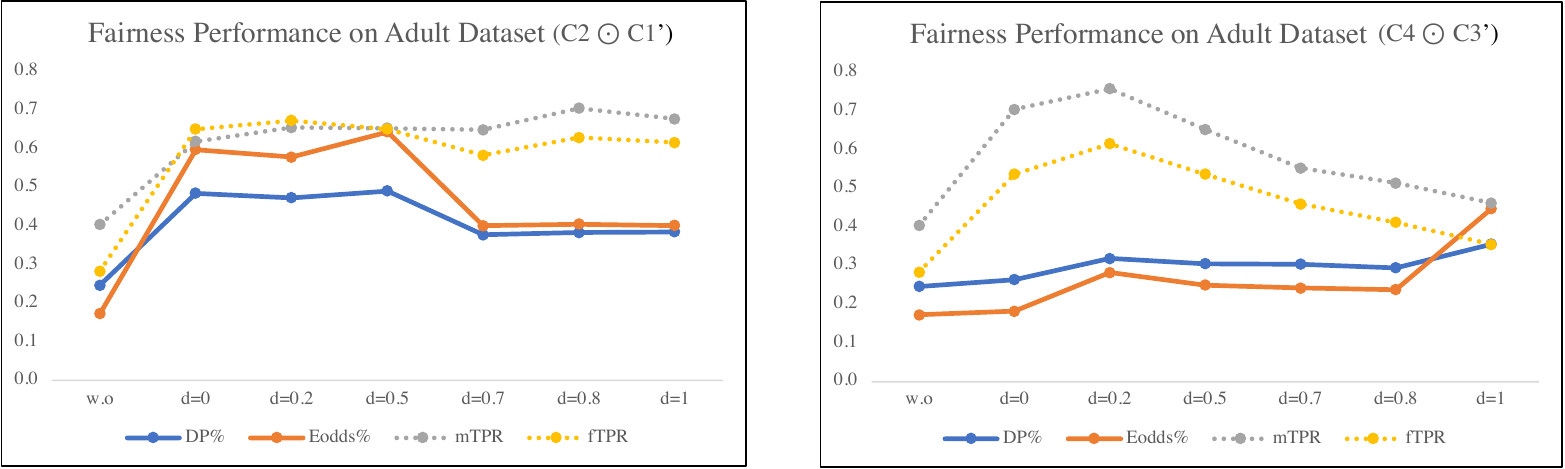}
    \caption{The fairness performance changes under different balancing degree $d$ in the Adult dataset under MLP model (fTPR: TPR in female group, mTPF: TPR in male group, \textit{d = [0, 0.2, 0.5, 0.7, 0.8, 1]}, refer to Appendix \ref{ap:adult_appendix} for detailed results)}
    \label{fig:mixed-demo-adult}
\end{figure}

We noticed the best fairness DP\% and Eodds\% occurs at $d=1$ under $C4 \odot C3'$. However, both TPR of female and male groups decline when $d$ exceeds \textit{0.5}. \cite{maheshwari2023fair} mentions a similar scenario and suggests to consider both relative and absolute values in fairness performance.
To have a further investigation of their performance in absolute values, Table \ref{tab:adult-split-c4} presents the model's performance across different subgroups. We can see the model trained with data augmentation in the \textit{0} to \textit{0.5} range, although having lower fairness metrics compared to $d=1$, shows an absolute improvement in model performance.
Therefore, we conclude the optimal balancing $d$ for $C4 \odot C3' $ strategy is \textit{0.2}.

\begin{table*}[t]
\centering
\caption{Subgroup-Level Performance Comparison on the Adult Dataset using $C4\odot C3'$ Sample (F-score refers performance on whole dataset, mF-score presents the F-score in male group, fF-score is the F-score in female group)}
\label{tab:adult-split-c4}
\setlength{\tabcolsep}{3.7mm}
\resizebox{0.98\linewidth}{!}{
\begin{tabular}{cccccccc}
\hline
 & F1 score & mF1 score & fF1 score & mTPR & fTPR & DP\% & Eodds\% \\
\hline
Baseline & 0.7003 & 0.6895 & 0.6833 & 0.4038 & 0.2831 & 0.2464 & 0.1730 \\ \hline
d=0 & \textbf{0.7976} & \textbf{0.7851} & \textbf{0.7898} & \underline{0.7046} & \underline{0.5368} & 0.2637 & 0.1822 \\
d=0.2 & \underline{0.7857} & \underline{0.7719} & 0.7776 & \textbf{0.7579} & \textbf{0.6158} & \underline{0.3188} & \underline{0.2821} \\
d=0.5 & 0.7841 & 0.7717 & \underline{0.7804} & 0.6523 & \underline{0.5368} & 0.3052 & 0.2499 \\
d=0.7 & 0.7651 & 0.7531 & 0.7633 & 0.5527 & 0.4596 & 0.3041 & 0.2420 \\
d=0.8 & 0.7529 & 0.7421 & 0.7453 & 0.5135 & 0.4118 & 0.2946 & 0.2381 \\
d=1 & 0.7198 & 0.7132 & 0.6927 & 0.4619 & 0.3548 & \textbf{0.3558} & \textbf{0.4471} \\ \hline
\end{tabular}
}
\end{table*}

\subsection{Counterfactual Cost across Different Groups}
\label{sec:res_cf}

We now evaluate the effectiveness of our algorithm from the XAI perspective, and the results are consistent with the above observations. 
First, we calculate the average (avg) and standard deviation (std) of the counterfactual cost across female (F) and male (M) subgroups. Then, we compare the cost gaps between the two groups. A smaller gap indicates fairer counterfactual explanations within different groups.
In the Adult dataset, $C2 \odot C1' $ remains to show more significant bias mitigation performance.
In the Law school dataset, as we disscussed above, the improvment is limited because the bias in the original dataset is not significant.

\begin{table*}[t]
\centering
\caption{Counterfactual explanations cost comparison on the Adult dataset with Decision Tree across female(F) and male(M) subgroups with different balancing degree $d=[0,0.5,1]$. }
\label{tab:res-diff-combinations}
\setlength{\tabcolsep}{1mm}
\resizebox{0.98\linewidth}{!}{
\begin{tabular}{ccccccccccccccc}
\hline
Strategy & Baseline     & \multicolumn{3}{c}{$C1 \odot C1'$}   & \multicolumn{3}{c}{$C2 \odot C1'$}   & \multicolumn{3}{c}{$C3 \odot C3'$}   & \multicolumn{3}{c}{$C4 \odot C3'$}   \\ 
\hline
$d$     & N/A    & 0      & 0.5    & 1      & 0      & 0.5    & 1      & 0      & 0.5    & 1      & 0      & 0.5    & 1      \\
\hline
$M_{avg}$ & 1.0049 & 1.5920 & 1.2988 & 1.2401 & 0.4484 & 0.7440 & 0.7817 & 1.1813 & 1.2762 & 1.1320 & 1.0904 & 0.5551 & 0.8681 \\
$M_{std}$ & 0.7893 & 1.1671 & 1.1786 & 1.1822 & 0.5500 & 0.4778 & 0.9267 & 1.2770 & 1.0981 & 1.1641 & 0.9260 & 0.8140 & 1.2145 \\
$F_{avg}$ & 1.0550 & 1.6986 & 1.3794 & 1.3447 & 0.4291 & 0.7738 & 0.7821 & 1.2001 & 1.3633 & 1.1906 & 1.0666 & 0.5221 & 0.7813 \\
$F_{std}$ & 0.8887 & 1.3167 & 1.2271 & 1.3758 & 0.5715 & 0.5184 & 1.0165 & 1.5087 & 1.3018 & 1.2127 & 0.9648 & 0.7750 & 1.1617 \\ \hline
$\Delta$avg & 0.0500 & 0.1066 & 0.0806 & 0.1047 & 0.0193 & 0.0298 & \textbf{0.0004} & \underline{0.0187} & 0.0871 & 0.0586 & 0.0238 & 0.0330 & 0.0868 \\
$\Delta$ std & 0.0994 & 0.1497 & 0.0485 & 0.1936 & \textbf{0.0215} & 0.0406 & 0.0899 & 0.2317 & \underline{0.2036} & 0.0486 & 0.0387 & 0.0391 & 0.0528 \\ 
\hline
\end{tabular}
}
\end{table*}

\begin{table*}[t]
\centering
\caption{Counterfactual Explanations cost comparison on Law dataset with Decision Tree across female(F) and male(M) subgroups with different balancing degree $d$}
\label{tab:res-cf-diff-combinations}
\setlength{\tabcolsep}{1mm}
\resizebox{0.98\linewidth}{!}{
\begin{tabular}{ccccccccccccccc}
\hline
Strategy & Baseline     & \multicolumn{3}{c}{$C1 \odot C1'$}   & \multicolumn{3}{c}{$C2 \odot C1'$}   & \multicolumn{3}{c}{$C3 \odot C3'$}   & \multicolumn{3}{c}{$C4 \odot C3'$}   \\ \hline
$d$ & N/A & 0 & 0.5 & 1 & 0 & 0.5 & 1 & 0 & 0.5 & 1 & 0 & 0.5 & 1 \\ \hline
$M_{avg}$ & 0.8671 & 0.8168 & 0.7392 & 0.9506 & 1.2589 & 0.7909 & 0.8026 & 0.8089 & 0.8415 & 0.6298 & 1.0218 & 0.8243 & 0.6968 \\
$M_{std}$ & 0.9075 & 0.7896 & 0.8382 & 0.9877 & 1.2913 & 0.8120 & 0.8901 & 0.8039 & 0.9753 & 0.6942 & 1.0415 & 0.9347 & 0.6902 \\
$F_{avg}$ & 0.9289 & 0.9444 & 0.8134 & 1.0363 & 1.4756 & 0.8351 & 0.8688 & 0.9384 & 0.9278 & 0.7033 & 1.1637 & 0.8763 & 0.6768 \\
$F_{std}$ & 0.9929 & 0.9233 & 0.8645 & 1.0259 & 1.5175 & 0.8293 & 0.9724 & 0.9269 & 1.0245 & 0.7914 & 1.1162 & 1.0108 & 0.6904 \\ \hline
$\Delta$avg & 0.0618 & 0.1276 & 0.0742 & 0.0857 & 0.2167 & \underline{0.0442} & 0.0662 & 0.1295 & 0.0863 & 0.0735 & 0.1419 & 0.0519 & \textbf{0.0200} \\
$\Delta$ std & 0.0854 & 0.1336 & 0.0263 & 0.0382 & 0.2262 & \underline{0.0173} & 0.0823 & 0.1230 & 0.0492 & 0.0971 & 0.0747 & 0.0761 & \textbf{0.0001}\\ \hline
\end{tabular}
}
\end{table*}

\section{Conclusion}

This paper proposed a new debiasing algorithm called ProxiMix. It extends the mixup technique by considering labels from proximity samples in the subgroup to mitigate potential bias in the preprocessing stage.
Our experiments evaluated the performance of ProxiMix with different sampling combinations and balancing degrees. The results prove that adding proximity-based labels improves fairness performance, and there exists optimal balancing degree for achieving the most significant enhancement. These observations were further supported by the experimental results on the cost comparison of counterfactual explanations. In future work, we plan to extent ProxiMix to multi-class tasks and consider intersectional fairness.

\begin{acknowledgments}
This work is funded by Doctoral Training Partnership Studentship of Engineering and Physical Sciences Research Council (EPSRC-DTP, EP/W524414/1/2894964).

\end{acknowledgments}

\appendix

\bibliography{ref.bib}

\begin{thebibliography}{37}
\expandafter\ifx\csname natexlab\endcsname\relax\def\natexlab#1{#1}\fi
\providecommand{\url}[1]{\texttt{#1}}
\providecommand{\href}[2]{#2}
\providecommand{\path}[1]{#1}
\providecommand{\DOIprefix}{doi:}
\providecommand{\ArXivprefix}{arXiv:}
\providecommand{\URLprefix}{URL: }
\providecommand{\Pubmedprefix}{pmid:}
\providecommand{\doi}[1]{\href{http://dx.doi.org/#1}{\path{#1}}}
\providecommand{\Pubmed}[1]{\href{pmid:#1}{\path{#1}}}
\providecommand{\bibinfo}[2]{#2}
\ifx\xfnm\relax \def\xfnm[#1]{\unskip,\space#1}\fi
\bibitem[{Osoba et~al.(2017)Osoba, Welser~IV, and Welser}]{osoba2017intelligence}
\bibinfo{author}{O.~A. Osoba}, \bibinfo{author}{W.~Welser~IV}, \bibinfo{author}{W.~Welser}, \bibinfo{title}{An intelligence in our image: The risks of bias and errors in artificial intelligence}, \bibinfo{publisher}{Rand Corporation}, \bibinfo{year}{2017}.
\bibitem[{Burkart and Huber(2021)}]{burkart2021survey}
\bibinfo{author}{N.~Burkart}, \bibinfo{author}{M.~F. Huber},
\newblock \bibinfo{title}{A survey on the explainability of supervised machine learning},
\newblock \bibinfo{journal}{Journal of Artificial Intelligence Research} \bibinfo{volume}{70} (\bibinfo{year}{2021}) \bibinfo{pages}{245--317}.
\bibitem[{Arrieta et~al.(2020)Arrieta, D{\'\i}az-Rodr{\'\i}guez, Del~Ser, Bennetot, Tabik, Barbado, Garc{\'\i}a, Gil-L{\'o}pez, Molina, Benjamins et~al.}]{arrieta2020explainable}
\bibinfo{author}{A.~B. Arrieta}, \bibinfo{author}{N.~D{\'\i}az-Rodr{\'\i}guez}, \bibinfo{author}{J.~Del~Ser}, \bibinfo{author}{A.~Bennetot}, \bibinfo{author}{S.~Tabik}, \bibinfo{author}{A.~Barbado}, \bibinfo{author}{S.~Garc{\'\i}a}, \bibinfo{author}{S.~Gil-L{\'o}pez}, \bibinfo{author}{D.~Molina}, \bibinfo{author}{R.~Benjamins}, et~al.,
\newblock \bibinfo{title}{Explainable artificial intelligence (xai): Concepts, taxonomies, opportunities and challenges toward responsible ai},
\newblock \bibinfo{journal}{Information fusion} \bibinfo{volume}{58} (\bibinfo{year}{2020}) \bibinfo{pages}{82--115}.
\bibitem[{Ciatto et~al.(2024)Ciatto, Sabbatini, Agiollo, Magnini, and Omicini}]{ciatto2024symbolic}
\bibinfo{author}{G.~Ciatto}, \bibinfo{author}{F.~Sabbatini}, \bibinfo{author}{A.~Agiollo}, \bibinfo{author}{M.~Magnini}, \bibinfo{author}{A.~Omicini},
\newblock \bibinfo{title}{Symbolic knowledge extraction and injection with sub-symbolic predictors: A systematic literature review},
\newblock \bibinfo{journal}{ACM Computing Surveys} \bibinfo{volume}{56} (\bibinfo{year}{2024}) \bibinfo{pages}{1--35}.
\bibitem[{Raji and Buolamwini(2019)}]{raji2019actionable}
\bibinfo{author}{I.~D. Raji}, \bibinfo{author}{J.~Buolamwini},
\newblock \bibinfo{title}{Actionable auditing: Investigating the impact of publicly naming biased performance results of commercial ai products},
\newblock in: \bibinfo{booktitle}{Proceedings of the 2019 AAAI/ACM Conference on AI, Ethics, and Society}, \bibinfo{year}{2019}, pp. \bibinfo{pages}{429--435}.
\bibitem[{Kavouras et~al.(2023)Kavouras, Tsopelas, Giannopoulos, Sacharidis, Psaroudaki, Theologitis, Rontogiannis, Fotakis, and Emiris}]{kavouras2023fairness}
\bibinfo{author}{L.~Kavouras}, \bibinfo{author}{K.~Tsopelas}, \bibinfo{author}{G.~Giannopoulos}, \bibinfo{author}{D.~Sacharidis}, \bibinfo{author}{E.~Psaroudaki}, \bibinfo{author}{N.~Theologitis}, \bibinfo{author}{D.~Rontogiannis}, \bibinfo{author}{D.~Fotakis}, \bibinfo{author}{I.~Emiris},
\newblock \bibinfo{title}{Fairness aware counterfactuals for subgroups},
\newblock in: \bibinfo{booktitle}{Thirty-seventh Conference on Neural Information Processing Systems}, \bibinfo{year}{2023}.
\bibitem[{Gohar and Cheng(2023)}]{gohar2023survey}
\bibinfo{author}{U.~Gohar}, \bibinfo{author}{L.~Cheng},
\newblock \bibinfo{title}{A survey on intersectional fairness in machine learning: Notions, mitigation, and challenges},
\newblock \bibinfo{journal}{arXiv preprint arXiv:2305.06969}  (\bibinfo{year}{2023}).
\bibitem[{Hort et~al.(2022)Hort, Chen, Zhang, Sarro, and Harman}]{hort2022bia}
\bibinfo{author}{M.~Hort}, \bibinfo{author}{Z.~Chen}, \bibinfo{author}{J.~M. Zhang}, \bibinfo{author}{F.~Sarro}, \bibinfo{author}{M.~Harman},
\newblock \bibinfo{title}{Bia mitigation for machine learning classifiers: A comprehensive survey},
\newblock \bibinfo{journal}{arXiv preprint arXiv:2207.07068}  (\bibinfo{year}{2022}).
\bibitem[{Zhang et~al.(2017)Zhang, Cisse, Dauphin, and Lopez-Paz}]{zhang2017mixup}
\bibinfo{author}{H.~Zhang}, \bibinfo{author}{M.~Cisse}, \bibinfo{author}{Y.~N. Dauphin}, \bibinfo{author}{D.~Lopez-Paz},
\newblock \bibinfo{title}{mixup: Beyond empirical risk minimization},
\newblock \bibinfo{journal}{arXiv preprint arXiv:1710.09412}  (\bibinfo{year}{2017}).
\bibitem[{Navarro et~al.(2023)Navarro, Little, Allen, and Segarra}]{navarro2023data}
\bibinfo{author}{M.~Navarro}, \bibinfo{author}{C.~Little}, \bibinfo{author}{G.~I. Allen}, \bibinfo{author}{S.~Segarra},
\newblock \bibinfo{title}{Data augmentation via subgroup mixup for improving fairness},
\newblock \bibinfo{journal}{arXiv preprint arXiv:2309.07110}  (\bibinfo{year}{2023}).
\bibitem[{Chuang and Mroueh(2021)}]{chuang2021fair}
\bibinfo{author}{C.-Y. Chuang}, \bibinfo{author}{Y.~Mroueh},
\newblock \bibinfo{title}{Fair mixup: Fairness via interpolation},
\newblock \bibinfo{journal}{arXiv preprint arXiv:2103.06503}  (\bibinfo{year}{2021}).
\bibitem[{Luong et~al.(2011)Luong, Ruggieri, and Turini}]{luong2011k}
\bibinfo{author}{B.~T. Luong}, \bibinfo{author}{S.~Ruggieri}, \bibinfo{author}{F.~Turini},
\newblock \bibinfo{title}{K-nn as an implementation of situation testing for discrimination discovery and prevention},
\newblock in: \bibinfo{booktitle}{Proceedings of the 17th ACM SIGKDD international conference on Knowledge discovery and data mining}, \bibinfo{year}{2011}, pp. \bibinfo{pages}{502--510}.
\bibitem[{Zafar et~al.(2017)Zafar, Valera, Gomez~Rodriguez, and Gummadi}]{zafar2017fairness}
\bibinfo{author}{M.~B. Zafar}, \bibinfo{author}{I.~Valera}, \bibinfo{author}{M.~Gomez~Rodriguez}, \bibinfo{author}{K.~P. Gummadi},
\newblock \bibinfo{title}{Fairness beyond disparate treatment \& disparate impact: Learning classification without disparate mistreatment},
\newblock in: \bibinfo{booktitle}{Proceedings of the 26th international conference on world wide web}, \bibinfo{year}{2017}, pp. \bibinfo{pages}{1171--1180}.
\bibitem[{Hardt et~al.(2016)Hardt, Price, and Srebro}]{hardt2016equality}
\bibinfo{author}{M.~Hardt}, \bibinfo{author}{E.~Price}, \bibinfo{author}{N.~Srebro},
\newblock \bibinfo{title}{Equality of opportunity in supervised learning},
\newblock \bibinfo{journal}{Advances in neural information processing systems} \bibinfo{volume}{29} (\bibinfo{year}{2016}).
\bibitem[{Buchanan and Shortliffe(1984)}]{buchanan1984rule}
\bibinfo{author}{B.~G. Buchanan}, \bibinfo{author}{E.~H. Shortliffe}, \bibinfo{title}{Rule based expert systems: the mycin experiments of the stanford heuristic programming project (the Addison-Wesley series in artificial intelligence)}, \bibinfo{publisher}{Addison-Wesley Longman Publishing Co., Inc.}, \bibinfo{year}{1984}.
\bibitem[{Gregor and Benbasat(1999)}]{gregor1999explanations}
\bibinfo{author}{S.~Gregor}, \bibinfo{author}{I.~Benbasat},
\newblock \bibinfo{title}{Explanations from intelligent systems: Theoretical foundations and implications for practice},
\newblock \bibinfo{journal}{MIS quarterly}  (\bibinfo{year}{1999}) \bibinfo{pages}{497--530}.
\bibitem[{Mothilal et~al.(2020)Mothilal, Sharma, and Tan}]{mothilal2020explaining}
\bibinfo{author}{R.~K. Mothilal}, \bibinfo{author}{A.~Sharma}, \bibinfo{author}{C.~Tan},
\newblock \bibinfo{title}{Explaining machine learning classifiers through diverse counterfactual explanations},
\newblock in: \bibinfo{booktitle}{Proceedings of the 2020 conference on fairness, accountability, and transparency}, \bibinfo{year}{2020}, pp. \bibinfo{pages}{607--617}.
\bibitem[{Wachter et~al.(2017)Wachter, Mittelstadt, and Russell}]{wachter2017counterfactual}
\bibinfo{author}{S.~Wachter}, \bibinfo{author}{B.~Mittelstadt}, \bibinfo{author}{C.~Russell},
\newblock \bibinfo{title}{Counterfactual explanations without opening the black box: Automated decisions and the gdpr},
\newblock \bibinfo{journal}{Harv. JL Tech.} \bibinfo{volume}{31} (\bibinfo{year}{2017}) \bibinfo{pages}{841}.
\bibitem[{Brughmans et~al.(2023)Brughmans, Leyman, and Martens}]{brughmans2023nice}
\bibinfo{author}{D.~Brughmans}, \bibinfo{author}{P.~Leyman}, \bibinfo{author}{D.~Martens},
\newblock \bibinfo{title}{Nice: an algorithm for nearest instance counterfactual explanations},
\newblock \bibinfo{journal}{Data Mining and Knowledge Discovery}  (\bibinfo{year}{2023}) \bibinfo{pages}{1--39}.
\bibitem[{Gupta et~al.(2019)Gupta, Nokhiz, Roy, and Venkatasubramanian}]{gupta2019equalizing}
\bibinfo{author}{V.~Gupta}, \bibinfo{author}{P.~Nokhiz}, \bibinfo{author}{C.~D. Roy}, \bibinfo{author}{S.~Venkatasubramanian},
\newblock \bibinfo{title}{Equalizing recourse across groups},
\newblock \bibinfo{journal}{arXiv preprint arXiv:1909.03166}  (\bibinfo{year}{2019}).
\bibitem[{Artelt and Hammer(2022)}]{artelt2022explain}
\bibinfo{author}{A.~Artelt}, \bibinfo{author}{B.~Hammer},
\newblock \bibinfo{title}{Explain it in the same way!--model-agnostic group fairness of counterfactual explanations},
\newblock \bibinfo{journal}{arXiv preprint arXiv:2211.14858}  (\bibinfo{year}{2022}).
\bibitem[{Goethals et~al.(2023)Goethals, Martens, and Calders}]{goethals2023precof}
\bibinfo{author}{S.~Goethals}, \bibinfo{author}{D.~Martens}, \bibinfo{author}{T.~Calders},
\newblock \bibinfo{title}{Precof: counterfactual explanations for fairness},
\newblock \bibinfo{journal}{Machine Learning}  (\bibinfo{year}{2023}) \bibinfo{pages}{1--32}.
\bibitem[{Friedler et~al.(2019)Friedler, Scheidegger, Venkatasubramanian, Choudhary, Hamilton, and Roth}]{friedler2019comparative}
\bibinfo{author}{S.~A. Friedler}, \bibinfo{author}{C.~Scheidegger}, \bibinfo{author}{S.~Venkatasubramanian}, \bibinfo{author}{S.~Choudhary}, \bibinfo{author}{E.~P. Hamilton}, \bibinfo{author}{D.~Roth},
\newblock \bibinfo{title}{A comparative study of fairness-enhancing interventions in machine learning},
\newblock in: \bibinfo{booktitle}{Proceedings of the conference on fairness, accountability, and transparency}, \bibinfo{year}{2019}, pp. \bibinfo{pages}{329--338}.
\bibitem[{Kamiran and Calders(2009)}]{kamiran2009classifying}
\bibinfo{author}{F.~Kamiran}, \bibinfo{author}{T.~Calders},
\newblock \bibinfo{title}{Classifying without discriminating},
\newblock in: \bibinfo{booktitle}{2009 2nd international conference on computer, control and communication}, \bibinfo{organization}{IEEE}, \bibinfo{year}{2009}, pp. \bibinfo{pages}{1--6}.
\bibitem[{Feldman et~al.(2015)Feldman, Friedler, Moeller, Scheidegger, and Venkatasubramanian}]{feldman2015certifying}
\bibinfo{author}{M.~Feldman}, \bibinfo{author}{S.~A. Friedler}, \bibinfo{author}{J.~Moeller}, \bibinfo{author}{C.~Scheidegger}, \bibinfo{author}{S.~Venkatasubramanian},
\newblock \bibinfo{title}{Certifying and removing disparate impact},
\newblock in: \bibinfo{booktitle}{proceedings of the 21th ACM SIGKDD international conference on knowledge discovery and data mining}, \bibinfo{year}{2015}, pp. \bibinfo{pages}{259--268}.
\bibitem[{Sun et~al.(2022)Sun, Wu, Wang, and Wang}]{sun2022towards}
\bibinfo{author}{H.~Sun}, \bibinfo{author}{K.~Wu}, \bibinfo{author}{T.~Wang}, \bibinfo{author}{W.~H. Wang},
\newblock \bibinfo{title}{Towards fair and robust classification},
\newblock in: \bibinfo{booktitle}{2022 IEEE 7th European Symposium on Security and Privacy (EuroS\&P)}, \bibinfo{organization}{IEEE}, \bibinfo{year}{2022}, pp. \bibinfo{pages}{356--376}.
\bibitem[{Kamiran et~al.(2010)Kamiran, Calders, and Pechenizkiy}]{kamiran2010discrimination}
\bibinfo{author}{F.~Kamiran}, \bibinfo{author}{T.~Calders}, \bibinfo{author}{M.~Pechenizkiy},
\newblock \bibinfo{title}{Discrimination aware decision tree learning},
\newblock in: \bibinfo{booktitle}{2010 IEEE international conference on data mining}, \bibinfo{organization}{IEEE}, \bibinfo{year}{2010}, pp. \bibinfo{pages}{869--874}.
\bibitem[{Pleiss et~al.(2017)Pleiss, Raghavan, Wu, Kleinberg, and Weinberger}]{pleiss2017fairness}
\bibinfo{author}{G.~Pleiss}, \bibinfo{author}{M.~Raghavan}, \bibinfo{author}{F.~Wu}, \bibinfo{author}{J.~Kleinberg}, \bibinfo{author}{K.~Q. Weinberger},
\newblock \bibinfo{title}{On fairness and calibration},
\newblock \bibinfo{journal}{Advances in neural information processing systems} \bibinfo{volume}{30} (\bibinfo{year}{2017}).
\bibitem[{Amend and Spurlock(2021)}]{amend2021improving}
\bibinfo{author}{J.~J. Amend}, \bibinfo{author}{S.~Spurlock},
\newblock \bibinfo{title}{Improving machine learning fairness with sampling and adversarial learning},
\newblock \bibinfo{journal}{J. Comput. Sci. Coll} \bibinfo{volume}{36} (\bibinfo{year}{2021}) \bibinfo{pages}{14--23}.
\bibitem[{Morano(2020)}]{morano2020bias}
\bibinfo{author}{A.~Morano},
\newblock \bibinfo{title}{Bias mitigation for automated decision making systems},
\newblock \bibinfo{journal}{Politecnico di Torino}  (\bibinfo{year}{2020}).
\bibitem[{Dablain et~al.(2022)Dablain, Krawczyk, and Chawla}]{dablain2022towards}
\bibinfo{author}{D.~Dablain}, \bibinfo{author}{B.~Krawczyk}, \bibinfo{author}{N.~Chawla},
\newblock \bibinfo{title}{Towards a holistic view of bias in machine learning: Bridging algorithmic fairness and imbalanced learning},
\newblock \bibinfo{journal}{arXiv preprint arXiv:2207.06084}  (\bibinfo{year}{2022}).
\bibitem[{Chakraborty et~al.(2021)Chakraborty, Majumder, and Menzies}]{biasWhyHow222}
\bibinfo{author}{J.~Chakraborty}, \bibinfo{author}{S.~Majumder}, \bibinfo{author}{T.~Menzies},
\newblock \bibinfo{title}{Bias in machine learning software: Why? how? what to do?},
\newblock \bibinfo{journal}{CoRR}  (\bibinfo{year}{2021}).
\bibitem[{Kohavi et~al.(1996)}]{adultdsref}
\bibinfo{author}{R.~Kohavi}, et~al.,
\newblock \bibinfo{title}{Scaling up the accuracy of naive-bayes classifiers: A decision-tree hybrid.},
\newblock in: \bibinfo{booktitle}{Kdd}, volume~\bibinfo{volume}{96}, \bibinfo{year}{1996}, pp. \bibinfo{pages}{202--207}.
\bibitem[{Xivuri and Twinomurinzi(2021)}]{lawschoolref}
\bibinfo{author}{K.~Xivuri}, \bibinfo{author}{H.~Twinomurinzi},
\newblock \bibinfo{title}{A systematic review of fairness in artificial intelligence algorithms},
\newblock in: \bibinfo{booktitle}{Responsible AI and Analytics for an Ethical and Inclusive Digitized Society: 20th IFIP WG 6.11 Conference on e-Business, e-Services and e-Society, I3E 2021, Galway, Ireland, September 1--3, 2021, Proceedings 20}, \bibinfo{organization}{Springer}, \bibinfo{year}{2021}, pp. \bibinfo{pages}{271--284}.
\bibitem[{Yeh and Lien(2009)}]{creditdsref}
\bibinfo{author}{I.-C. Yeh}, \bibinfo{author}{C.-h. Lien},
\newblock \bibinfo{title}{The comparisons of data mining techniques for the predictive accuracy of probability of default of credit card clients},
\newblock \bibinfo{journal}{Expert systems with applications} \bibinfo{volume}{36} (\bibinfo{year}{2009}) \bibinfo{pages}{2473--2480}.
\bibitem[{Maheshwari et~al.(2023)Maheshwari, Bellet, Denis, and Keller}]{maheshwari2023fair}
\bibinfo{author}{G.~Maheshwari}, \bibinfo{author}{A.~Bellet}, \bibinfo{author}{P.~Denis}, \bibinfo{author}{M.~Keller},
\newblock \bibinfo{title}{Fair without leveling down: A new intersectional fairness definition},
\newblock in: \bibinfo{booktitle}{EMNLP 2023-The 2023 Conference on Empirical Methods in Natural Language Processing}, \bibinfo{year}{2023}.
\bibitem[{Le~Quy et~al.(2022)Le~Quy, Roy, Iosifidis, Zhang, and Ntoutsi}]{le2022survey}
\bibinfo{author}{T.~Le~Quy}, \bibinfo{author}{A.~Roy}, \bibinfo{author}{V.~Iosifidis}, \bibinfo{author}{W.~Zhang}, \bibinfo{author}{E.~Ntoutsi},
\newblock \bibinfo{title}{A survey on datasets for fairness-aware machine learning},
\newblock \bibinfo{journal}{Wiley Interdisciplinary Reviews: Data Mining and Knowledge Discovery} \bibinfo{volume}{12} (\bibinfo{year}{2022}) \bibinfo{pages}{e1452}.

\end{thebibliography}


\section{Appendices: Dataset Description}
\subsection{Adult Income Dataset} 

The Adult Income dataset is also known as the Census Income dataset.
Its documentation \footnote{\url{https://www.cs.toronto.edu/~delve/data/adult/adultDetail.html}} provides a detailed description of 14 features in the dataset. We omitted some features, such as `fnlwgt', and the final features we used after data cleaning are as follows.

\begin{table*}[h]
\caption{The Adult Dataset Descriptions}
\label{tab:adult-ds}
\begin{tabular}{ccc}
\hline
Feature Name & Value Type & Description \\ \hline
Sex (sensitive features) & Categories & Gender of the persom, eg. Male, Female \\
Workclass & Categories & Type of employment, eg. Private,  Self-employed \\
Age & Continuous & Age of the person \\
Education & Categories & Highest level of education, eg. Bachelors, Some-college \\
Education-num & Continuous & Education level of the person \\
Marital-Status & Categories & Marital status of the persom, eg. Single, Married \\
Occupation & Categories & Occupation of the persom, eg. Tech-support, Sales \\
Relationship & Categories & Role in the family, eg. Not-in-family, Own-child \\
Capital-Gain & Continuous & Capital gains of the persom \\
Capital-Loss & Continuous & Capital loss  of the persom \\
Hours-Per-Week & Continuous & Hours worked per week \\
Race & Categories & Race of the person, eg. White, Other \\ 
Salary (ground truth Y) & Categories & Whether annual income exceeds $50K$ \\

\hline
\end{tabular}
\end{table*}

\newpage
\subsection{Law School Dataset}
The Law School dataset contains admission records for law schools. We followed the description provided in \cite{le2022survey} and the data cleaning pipeline in \cite{artelt2022explain}, extracting the following features for the experiment.
\begin{table*}[h]
\caption{The Law School Dataset Descriptions}
\label{tab:law-ds}
\begin{tabular}{ccc}
\hline
Feature Name & Value Types & Description \\
\hline
gender (sensitive feature) & Categories & Gender \\
race & Categories & Race \\
decile1 & Continuous & The decile in the school given his grades in Year 1 \\
decile3 & Continuous & The decile in the school given his grades in Year 3 \\
lsat & Continuous & LSAT score \\
ugpa & Continuous & Undergraduate GPA. \\
zfygpa & Continuous & The first year Law school GPA \\
zgpa & Continuous & The cumulative law school GPA. \\
fulltime & Categories & Work full-time or part-time \\
fam\_inc & Continuous & Family income \\
pass\_bar (ground truth Y) & Categories & Whether passed the bar exam.\\
\hline
\end{tabular}
\end{table*}

\subsection{Credit Default Dataset}

The Credit Default dataset, also known as the credit card clients dataset, explores default payments on credit cards. Followings are the features and descriptions.

\begin{table*}[ht]
\caption{The Credit Default Dataset Descriptions}
\begin{tabular}{ccc}
\hline
Attribute & Value Types & Description \\
\hline
SEX(sensitive feature) & Categories & Gender \\
EDUCATION & Categories & Highest education \\
AGE & Continuous & Age \\
LIMIT\_BAL & Continuous & Amount of given credit \\
PAY\_i ($i \in \{1,2,3,4,5,6 \}$) & Continuous & Repayment status for $i$th month \\
BILL\_AMT\_i ($i \in \{1,2,3,4,5,6 \}$) & Continuous & Amount of bill statement for $i$th month \\
PAY\_AMT\_i ($i \in \{1,2,3,4,5,6 \}$) & Continuous & Amount of previous payment for $i$th month \\
Default\_Payment(ground truth Y) & Categories & Whether default payment or not next month\\
\hline
\end{tabular}
\end{table*}

\section{Appendices: Results}
\label{ap:table}

\newpage
\subsection{ProxiMix in Credit Default Dataset with MLP model}
\label{app:credit_appendix}

\begin{table*}[h]
\caption{The prediction and fairness performance under different balancing degree $d$ in Credit Default dataset with MLP model (Acc: accuracy, F1: F1-score, m: performance in male subgroup, f: performance in female subgroup)}
\resizebox{\linewidth}{!}{
\begin{tabular}{cccccccccccccc}
\hline
d & Strategy & Acc. & F1 & $\Delta DP$ & DP\% & $\Delta Eodds$ & Eodds\% & mF1 & mTPR & mFPR & fF1 & fTPR & fFPR \\
\hline
d=0 & $C1 \odot C1'$ & 0.7829 & 0.5075 & 0.0170 & 0.5885 & 0.0278 & 0.5263 & 0.5131 & 0.0953 & 0.0243 & 0.5022 & 0.0675 & 0.0128 \\
d=0.2 & $C1 \odot C1'$ & 0.7802 & 0.4906 & 0.0147 & 0.5625 & 0.0259 & 0.5318 & 0.4964 & 0.0741 & 0.0210 & 0.4853 & 0.0482 & 0.0112 \\
d=0.5 & $C1 \odot C1'$ & 0.7708 & 0.5217 & 0.0119 & 0.8156 & 0.0239 & 0.8082 & 0.5266 & 0.1247 & 0.0453 & 0.5174 & 0.1008 & 0.0396 \\
d=0.7 & $C1 \odot C1'$ & 0.6877 & 0.5393 & 0.0181 & 0.9171 & 0.0205 & 0.8994 & 0.5404 & 0.2600 & 0.1817 & 0.5386 & 0.2805 & 0.2020 \\
d=0.8 & $C1 \odot C1'$ & 0.7574 & 0.5922 & 0.0303 & 0.8111 & 0.0289 & 0.7643 & 0.5853 & 0.2812 & 0.1227 & 0.5963 & 0.2673 & 0.0938 \\
d=1 & $C1 \odot C1'$ & 0.7569 & 0.6006 & 0.0446 & 0.7523 & 0.0463 & 0.6712 & 0.5880 & 0.3059 & 0.1408 & 0.6087 & 0.2901 & 0.0945 \\
\hline
 & Baseline & 0.7781 & 0.5076 & 0.0233 & 0.5485 & 0.0352 & 0.5000 & 0.5147 & 0.1035 & 0.0354 & 0.5008 & 0.0684 & 0.0177\\
\hline
d=0 & $C2 \odot C1'$ & 0.5277 & 0.5013 & 0.0623 & 0.8937 & 0.0661 & 0.8807 & 0.4884 & 0.6894 & 0.5540 & 0.5083 & 0.6599 & 0.4879 \\
d=0.2 & $C2 \odot C1'$ & 0.4959 & 0.4775 & 0.0358 & 0.9408 & 0.0477 & 0.9177 & 0.5026 & 0.6894 & 0.5319 & 0.4611 & 0.7020 & 0.5796 \\
d=0.5 & $C2 \odot C1'$ & 0.5599 & 0.5262 & 0.0080 & 0.9844 & 0.0145 & 0.9695 & 0.5376 & 0.6565 & 0.4604 & 0.5185 & 0.6670 & 0.4749 \\
d=0.7 & $C2 \odot C1'$ & 0.6998 & 0.6079 & 0.0333 & 0.8942 & 0.0306 & 0.8816 & 0.6032 & 0.4941 & 0.2588 & 0.6105 & 0.4829 & 0.2281 \\
d=0.8 & $C2 \odot C1'$ & 0.7047 & 0.5520 & 0.0134 & 0.9332 & 0.0173 & 0.9037 & 0.5542 & 0.2659 & 0.1622 & 0.5504 & 0.2787 & 0.1795 \\
d=1 & $C2 \odot C1'$ & 0.6466 & 0.5563 & 0.0200 & 0.9410 & 0.0179 & 0.9416 & 0.5559 & 0.4471 & 0.3063 & 0.5560 & 0.4382 & 0.2884 \\
\hline
 & Baseline & 0.7781 & 0.5076 & 0.0233 & 0.5485 & 0.0352 & 0.5000 & 0.5147 & 0.1035 & 0.0354 & 0.5008 & 0.0684 & 0.0177\\
\hline
d=0 & $C3 \odot C3'$ & 0.7757 & 0.4847 & 0.0114 & 0.6661 & 0.0194 & 0.6794 & 0.4877 & 0.0659 & 0.0243 & 0.4816 & 0.0465 & 0.0165 \\
d=0.2 & $C3 \odot C3'$ & 0.4284 & 0.4192 & 0.0322 & 0.9516 & 0.0460 & 0.9305 & 0.4478 & 0.6882 & 0.6156 & 0.4004 & 0.6784 & 0.6615 \\
d=0.5 & $C3 \odot C3'$ & 0.6440 & 0.5691 & 0.0495 & 0.8736 & 0.0492 & 0.8589 & 0.5618 & 0.5282 & 0.3491 & 0.5730 & 0.5022 & 0.2998 \\
d=0.7 & $C3 \odot C3'$ & 0.7641 & 0.5974 & 0.0371 & 0.7649 & 0.0389 & 0.6761 & 0.5849 & 0.2776 & 0.1202 & 0.6057 & 0.2691 & 0.0812 \\
d=0.8 & $C3 \odot C3'$ & 0.7203 & 0.5889 & 0.0339 & 0.8551 & 0.0352 & 0.8203 & 0.5795 & 0.3541 & 0.1961 & 0.5947 & 0.3471 & 0.1608 \\
d=1 & $C3 \odot C3'$ & 0.7822 & 0.4988 & 0.0179 & 0.5233 & 0.0225 & 0.3886 & 0.5013 & 0.0812 & 0.0240 & 0.4958 & 0.0587 & 0.0093 \\
\hline
 & Baseline & 0.7781 & 0.5076 & 0.0233 & 0.5485 & 0.0352 & 0.5000 & 0.5147 & 0.1035 & 0.0354 & 0.5008 & 0.0684 & 0.0177\\
\hline
d=0 & $C4 \odot C3'$ & 0.7749 & 0.5701 & 0.0379 & 0.6603 & 0.0361 & 0.5572 & 0.5650 & 0.2071 & 0.0815 & 0.5724 & 0.1797 & 0.0454 \\
d=0.2 & $C4 \odot C3'$ & 0.7747 & 0.5970 & 0.0460 & 0.6773 & 0.0460 & 0.6061 & 0.5961 & 0.2765 & 0.1006 & 0.5960 & 0.2305 & 0.0610 \\
d=0.5 & $C4 \odot C3'$ & 0.7844 & 0.5997 & 0.0310 & 0.7378 & 0.0330 & 0.6813 & 0.6002 & 0.2565 & 0.0748 & 0.5982 & 0.2235 & 0.0510 \\
d=0.7 & $C4 \odot C3'$ & 0.2873 & 0.2741 & 0.0112 & 0.9878 & 0.0149 & 0.9835 & 0.2935 & 0.9518 & 0.8931 & 0.2614 & 0.9571 & 0.9081 \\
d=0.8 & $C4 \odot C3'$ & 0.7806 & 0.5457 & 0.0302 & 0.6129 & 0.0344 & 0.5211 & 0.5482 & 0.1588 & 0.0527 & 0.5424 & 0.1245 & 0.0275 \\
d=1 & $C4 \odot C3'$ & 0.7824 & 0.4918 & 0.0166 & 0.4915 & 0.0259 & 0.3813 & 0.4970 & 0.0741 & 0.0195 & 0.4868 & 0.0482 & 0.0074 \\
\hline
 & Baseline & 0.7781 & 0.5076 & 0.0233 & 0.5485 & 0.0352 & 0.5000 & 0.5147 & 0.1035 & 0.0354 & 0.5008 & 0.0684 & 0.0177\\
\hline
\end{tabular}
}
\end{table*}

\newpage
\subsection{ProxiMix in Adult Income Dataset with MLP model}
\label{ap:adult_appendix}

\begin{table*}[ht]
\caption{The prediction and fairness performance under different balancing degree $d$ in the Adult dataset with MLP model (Acc: accuracy, F1: F1-score, m: performance in male subgroup, f: performance in female subgroup)}
\resizebox{\linewidth}{!}{
\begin{tabular}{ccccccccccccc}
\hline
d & Strategy & F1 & $\Delta DP$ & DP\% & $\Delta Eodds$ & Eodds\% & mF1 & mTPR & mFPR & fF1 & fTPR & fFPR \\
\hline
d=0 & $C1 \odot C1'$ & 0.7302 & 0.1653 & 0.1935 & 0.2128 & 0.1240 & 0.7238 & 0.4922 & 0.0793 & 0.6818 & 0.2794 & 0.0098 \\
d=0.2 & $C1 \odot C1'$ & 0.7834 & 0.1942 & 0.2905 & 0.1565 & 0.2571 & 0.7747 & 0.6455 & 0.1112 & 0.7618 & 0.4890 & 0.0286 \\
d=0.5 & $C1 \odot C1'$ & 0.7431 & 0.1427 & 0.2532 & 0.1374 & 0.1759 & 0.7342 & 0.4885 & 0.0611 & 0.7222 & 0.3511 & 0.0107 \\
d=0.7 & $C1 \odot C1'$ & 0.7823 & 0.1814 & 0.2864 & 0.1349 & 0.2291 & 0.7723 & 0.6165 & 0.0958 & 0.7693 & 0.4816 & 0.0220 \\
d=0.8 & $C1 \odot C1'$ & 0.7731 & 0.1637 & 0.2554 & 0.1547 & 0.1768 & 0.7652 & 0.5628 & 0.0698 & 0.7499 & 0.4081 & 0.0123 \\
d=1 & $C1 \odot C1'$ & 0.7723 & 0.2026 & 0.2385 & 0.1920 & 0.1635 & 0.7622 & 0.6185 & 0.1119 & 0.7492 & 0.4265 & 0.0183 \\
\hline
 & Baseline & 0.7003 & 0.1238 & 0.2464 & 0.1207 & 0.1730 & 0.6895 & 0.4038 & 0.0595 & 0.6833 & 0.2831 & 0.0103 \\
\hline
d=0 & $C2 \odot C1'$ & 0.7806 & 0.1325 & 0.4847 & 0.0398 & 0.5979 & 0.7711 & 0.6188 & 0.0991 & 0.7801 & 0.6507 & 0.0592 \\
d=0.2 & $C2 \odot C1'$ & 0.7901 & 0.1442 & 0.4729 & 0.0450 & 0.5785 & 0.7819 & 0.6550 & 0.1067 & 0.7843 & 0.6728 & 0.0617 \\
d=0.5 & $C2 \odot C1'$ & 0.7834 & 0.1402 & 0.4906 & 0.0392 & 0.6437 & 0.7787 & 0.6529 & 0.1101 & 0.7641 & 0.6507 & 0.0709 \\
d=0.7 & $C2 \odot C1'$ & 0.7886 & 0.1689 & 0.3773 & 0.0658 & 0.4007 & 0.7795 & 0.6485 & 0.1061 & 0.7793 & 0.5827 & 0.0425 \\
d=0.8 & $C2 \odot C1'$ & 0.7924 & 0.1902 & 0.3832 & 0.0804 & 0.4046 & 0.7837 & 0.7046 & 0.1351 & 0.7787 & 0.6287 & 0.0547 \\
d=1 & $C2 \odot C1'$ & 0.7891 & 0.1800 & 0.3846 & 0.0744 & 0.4011 & 0.7793 & 0.6769 & 0.1243 & 0.7809 & 0.6158 & 0.0499 \\
\hline
 & Baseline & 0.7003 & 0.1238 & 0.2464 & 0.1207 & 0.1730 & 0.6895 & 0.4038 & 0.0595 & 0.6833 & 0.2831 & 0.0103 \\
\hline
d=0 & $C3 \odot C3'$ & 0.7832 & 0.1997 & 0.3582 & 0.1149 & 0.3713 & 0.7747 & 0.6958 & 0.1429 & 0.7631 & 0.5809 & 0.0531 \\
d=0.2 & $C3 \odot C3'$ & 0.7626 & 0.1339 & 0.3412 & 0.0676 & 0.3369 & 0.7520 & 0.5253 & 0.0624 & 0.7597 & 0.4577 & 0.0210 \\
d=0.5 & $C3 \odot C3'$ & 0.7729 & 0.1646 & 0.2658 & 0.1466 & 0.1975 & 0.7645 & 0.5675 & 0.0741 & 0.7525 & 0.4210 & 0.0146 \\
d=0.7 & $C3 \odot C3'$ & 0.7911 & 0.2052 & 0.2884 & 0.1493 & 0.2360 & 0.7806 & 0.6732 & 0.1202 & 0.7775 & 0.5239 & 0.0284 \\
d=0.8 & $C3 \odot C3'$ & 0.7763 & 0.1912 & 0.2167 & 0.2198 & 0.1373 & 0.7700 & 0.6003 & 0.0883 & 0.7359 & 0.3805 & 0.0121 \\
d=1 & $C3 \odot C3'$ & 0.7573 & 0.1410 & 0.2634 & 0.1346 & 0.1936 & 0.7496 & 0.5078 & 0.0531 & 0.7350 & 0.3732 & 0.0103 \\
\hline
 & Baseline & 0.7003 & 0.1238 & 0.2464 & 0.1207 & 0.1730 & 0.6895 & 0.4038 & 0.0595 & 0.6833 & 0.2831 & 0.0103 \\
\hline
d=0 & $C4 \odot C3'$ & 0.7976 & 0.2260 & 0.2637 & 0.1678 & 0.1822 & 0.7851 & 0.7046 & 0.1330 & 0.7898 & 0.5368 & 0.0242 \\
d=0.2 & $C4 \odot C3'$ & 0.7857 & 0.2447 & 0.3188 & 0.1421 & 0.2821 & 0.7719 & 0.7579 & 0.1848 & 0.7776 & 0.6158 & 0.0521 \\
d=0.5 & $C4 \odot C3'$ & 0.7841 & 0.1958 & 0.3052 & 0.1155 & 0.2499 & 0.7717 & 0.6523 & 0.1199 & 0.7804 & 0.5368 & 0.0300 \\
d=0.7 & $C4 \odot C3'$ & 0.7651 & 0.1559 & 0.3041 & 0.0931 & 0.2420 & 0.7531 & 0.5527 & 0.0803 & 0.7633 & 0.4596 & 0.0194 \\
d=0.8 & $C4 \odot C3'$ & 0.7529 & 0.1437 & 0.2946 & 0.1017 & 0.2381 & 0.7421 & 0.5135 & 0.0682 & 0.7453 & 0.4118 & 0.0162 \\
d=1 & $C4 \odot C3'$ & 0.7198 & 0.1226 & 0.3558 & 0.1071 & 0.4471 & 0.7132 & 0.4619 & 0.0716 & 0.6927 & 0.3548 & 0.0320 \\
\hline
 & Baseline & 0.7003 & 0.1238 & 0.2464 & 0.1207 & 0.1730 & 0.6895 & 0.4038 & 0.0595 & 0.6833 & 0.2831 & 0.0103 \\
\hline
\end{tabular}
}
\end{table*}

\end{document}